\def\year{2022}\relax
\documentclass[letterpaper]{article} 
\usepackage{aaai22}  
\usepackage{times}  
\usepackage{helvet}  
\usepackage{courier}  
\usepackage[hyphens]{url}  
\usepackage{graphicx} 
\def\UrlFont{\rm}  
\usepackage{natbib}  
\usepackage{caption} 
\DeclareCaptionStyle{ruled}{labelfont=normalfont,labelsep=colon,strut=off} 
\usepackage{algorithm}
\usepackage[noend]{algorithmic}
\usepackage{amsmath}
\usepackage{amssymb}
\usepackage{booktabs}
\usepackage{multirow}

\usepackage{newfloat}
\usepackage{listings}
\floatstyle{ruled}
\newfloat{listing}{tb}{lst}{}
\floatname{listing}{Listing}
\title{TLogic: Temporal Logical Rules for Explainable Link Forecasting\\ on Temporal Knowledge Graphs}
\author{
Yushan Liu\textsuperscript{\rm 1,2},
    Yunpu Ma\textsuperscript{\rm 1,2},
     Marcel Hildebrandt\textsuperscript{\rm 1},
     Mitchell Joblin\textsuperscript{\rm 1},
     Volker Tresp\textsuperscript{\rm 1,2}
    
}
\affiliations{

\textsuperscript{\rm 1}Siemens AG, Otto-Hahn-Ring 6, 81739 Munich, Germany

\{firstname.lastname\}@siemens.com

\textsuperscript{\rm 2}Ludwig Maximilian University of Munich, Geschwister-Scholl-Platz 1, 80539 Munich, Germany

}

\begin{document}

\maketitle

\begin{abstract}
Conventional static knowledge graphs model entities in relational data as nodes, connected by edges of specific relation types.
However, information and knowledge evolve continuously, and temporal dynamics emerge, which are expected to influence future situations.
In temporal knowledge graphs, time information is integrated into the graph by equipping each edge with a timestamp or a time range.
Embedding-based methods have been introduced for link prediction on temporal knowledge graphs, but they mostly lack explainability and comprehensible reasoning chains. 
Particularly, they are usually not designed to deal with link forecasting -- event prediction involving future timestamps. 
We address the task of link forecasting on temporal knowledge graphs and introduce TLogic, an explainable framework that is based on temporal logical rules extracted via temporal random walks. 
We compare TLogic with state-of-the-art baselines on three benchmark datasets and show better overall performance while our method also provides explanations that preserve time consistency.
Furthermore, in contrast to most state-of-the-art embedding-based methods, TLogic works well in the inductive setting where already learned rules are transferred to related datasets with a common vocabulary.  
\end{abstract}

\section{Introduction}
Knowledge graphs (KGs) structure factual information in the form of triples $(e_s, r ,e_o)$, where $e_s$ and $e_o$ correspond to entities in the real world and $r$ to a binary relation, e.\,g., \textit{(Anna, born in, Paris)}. 
This knowledge representation leads to an interpretation as a directed multigraph, where entities are identified with nodes and relations with edge types. Each edge $(e_s, r ,e_o)$ in the KG encodes an observed fact, where the source node $e_s$ corresponds to the subject entity, the target node $e_o$ to the object entity, and the edge type $r$ to the predicate of the factual statement.

\begin{figure}[t]
\centering
\includegraphics[width=\columnwidth]{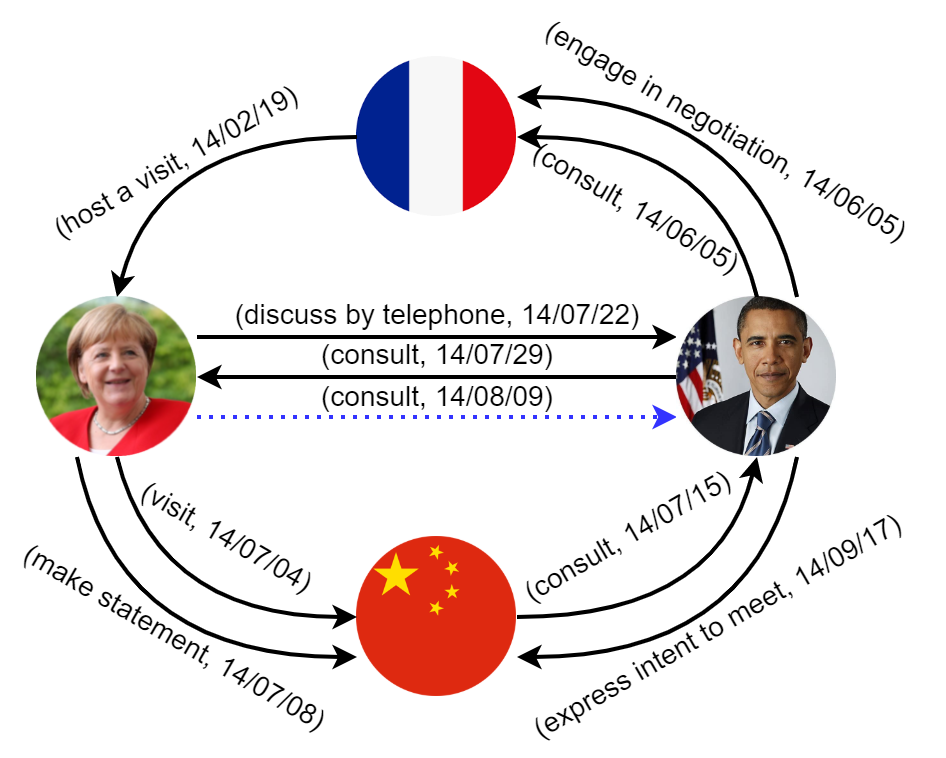}
\caption{A subgraph from the dataset ICEWS14 with the entities \textit{Angela Merkel, Barack Obama, France}, and \textit{China}. The timestamps are displayed in the format yy/mm/dd. The dotted blue line represents the correct answer to the query \textit{(Angela Merkel, consult, ?, 2014/08/09)}. Previous interactions between \textit{Angela Merkel} and \textit{Barack Obama} can be interpreted as an explanation for the prediction.}
\label{fig:tkg_example}
\end{figure}

Some real-world information also includes a temporal dimension, e.\,g., the event \textit{(Anna, born in, Paris)} happened on a specific date. To model the large amount of available event data that induce complex interactions between entities over time, temporal knowledge graphs (tKGs) have been introduced. Temporal KGs extend the triples to quadruples $(e_s, r, e_o, t)$ to integrate a timestamp or time range $t$, where $t$ indicates the time validity of the static event $(e_s, r, e_o)$, e.\,g., \mbox{\textit{(Angela Merkel, visit, China, 2014/07/04)}}. Figure~\ref{fig:tkg_example} visualizes a subgraph from the dataset ICEWS14 as an example of a tKG. In this work, we focus on tKGs where each edge is equipped with a single timestamp.

One of the common tasks on KGs is link prediction, which finds application in areas such as recommender systems~\cite{nectr.hildebrandt.2019}, knowledge base completion~\cite{convkb.nguyen.2018}, and drug repurposing~\cite{polo.liu.2021}. 
Taking the additional temporal dimension into account, it is of special interest to forecast events for future timestamps based on past information. Notable real-world applications that rely on accurate event forecasting are, e.\,g., clinical decision support, supply chain management, and extreme events modeling.
In this work, we address link forecasting on tKGs, where we consider queries $(e_s, r, ?, t)$ for a timestamp $t$ that has not been seen during training.

Several embedding-based methods have been introduced  for tKGs to solve link prediction and forecasting (link prediction with future timestamps), e.g., TTransE~\cite{ttranse.leblay.2018}, TNTComplEx~\cite{tntcomplex.lacroix.2020}, and RE-Net~\cite{renet.jin.2019}.
The underlying principle is to project the entities and relations into a low-dimensional vector space while preserving the topology and temporal dynamics of the tKG. These methods can learn the complex patterns that lead to an event but often lack transparency and interpretability.

To increase the transparency and trustworthiness of the solutions, human-understandable explanations are necessary, which can be provided by logical rules.
However, the manual creation of rules is often difficult due to the complex nature of events. Domain experts cannot articulate the conditions for the occurrence of an event sufficiently formally to express this knowledge as rules, which leads to a problem termed as the knowledge acquisition bottleneck. Generally, symbolic methods that make use of logical rules tend to suffer from scalability issues, which make them impractical for the application on large real-world datasets.

We propose TLogic that automatically mines cyclic temporal logical rules by extracting temporal random walks from the graph. We achieve both high predictive performance and time-consistent explanations in the form of temporal rules, which conform to the observation that the occurrence of an event is usually triggered by previous events. 
The main contributions of this work are summarized as follows:
\begin{itemize}
\item We introduce TLogic, a novel symbolic framework based on temporal random walks in temporal knowledge graphs. It is the first approach that directly learns temporal logical rules from tKGs and applies these rules to the link forecasting task. 
\item Our approach provides explicit and human-readable explanations in the form of temporal logical rules and is scalable to large datasets.
\item We conduct experiments on three benchmark datasets (ICEWS14, ICEWS18, and ICEWS0515) and show better overall performance compared with state-of-the-art baselines.
\item We demonstrate the effectiveness of our method in the inductive setting where our learned rules are transferred to a related dataset with a common vocabulary.
\end{itemize}

\section{Related Work}
Subsymbolic machine learning methods, e.\,g., embedding-based algorithms, have achieved success for the link prediction task on static KGs. Well-known methods include RESCAL~\cite{rescal.nickel.2011}, TransE~\cite{transe.bordes.2013}, DistMult~\cite{distmult.yang.2015}, and ComplEx~\cite{complex.trouillon.2016} as well as the graph convolutional approaches R-GCN~\cite{rgcn.schlichtkrull.2018} and CompGCN~\cite{compgcn.vashishth.2020}. 
Several approaches have been recently proposed to handle tKGs, such as TTransE~\cite{ttranse.leblay.2018}, TA-DistMult~\cite{ta-distmult-transe.garcia-duran.2018}, DE-SimplE~\cite{de-simple.goel.2020}, TNTComplEx~\cite{tntcomplex.lacroix.2020}, CyGNet~\cite{cygnet.zhu.2021}, RE-Net~\cite{renet.jin.2019}, and xERTE~\cite{xerte.han.2021}. The main idea of these methods is to explicitly learn embeddings for timestamps or to integrate temporal information into the entity or relation embeddings. 
However, the black-box property of embeddings makes it difficult for humans to understand the predictions. Moreover, approaches with shallow embeddings are not suitable for an inductive setting with previously unseen entities, relations, or timestamps. 
From the above methods, only CyGNet, RE-Net, and xERTE are designed for the forecasting task. xERTE is also able to provide explanations by extracting relevant subgraphs around the query subject. 

Symbolic approaches for link prediction on KGs like  AMIE+~\cite{amie+.galarraga.2015} and AnyBURL~\cite{anyburl.meilicke.2019} mine logical rules from the dataset, which are then applied to predict new links. StreamLearner~\cite{streamlearner.omran.2019} is one of the first methods for learning temporal rules. It employs a static rule learner to generate rules, which are then generalized to the temporal domain. However, they only consider a rather restricted set of temporal rules, where all body atoms have the same timestamp.

Another class of approaches is based on random walks in the graph, where the walks can support an interpretable explanation for the predictions. For example, AnyBURL samples random walks for generating rules. The methods dynnode2vec~\cite{dynnode2vec.mahdavi.2018} and change2vec~\cite{change2vec.bian.2019} alternately extract random walks on tKG snapshots and learn parameters for node embeddings, but they do not capture temporal patterns within the random walks. \citet{ctdne.nguyen.2018} extend the concept of random walks to temporal random walks on continuous-time dynamic networks for learning node embeddings, where the sequence of edges in the walk only moves forward in time.


\section{Preliminaries}
Let $[n] := \{1, 2, \dots, n\}$.

\paragraph{Temporal knowledge graph}
Let $\mathcal{E}$ denote the set of entities, $\mathcal{R}$ the set of relations, and $\mathcal{T}$ the set of timestamps.

A \textit{temporal knowledge graph} (tKG) is a collection of facts $\mathcal{G} \subset \mathcal{E} \times \mathcal{R} \times \mathcal{E} \times \mathcal{T}$, where each fact is represented by a quadruple $(e_s, r, e_o, t)$. 
The quadruple $(e_s, r, e_o, t)$ is also called link or edge, and it indicates a connection between the subject entity $e_s \in \mathcal{E}$ and the object entity $e_o \in \mathcal{E}$ via the relation $r \in \mathcal{R}$. 
The timestamp $t \in \mathcal{T}$ implies the occurrence of the event $(e_s, r, e_o)$ at time $t$, where $t$ can be measured in units such as hour, day, and year. 

For two timestamps $t$ and $\hat{t}$, we denote the fact that $t$ occurs earlier than $\hat{t}$ by $t < \hat{t}$. If additionally, $t$ could represent the same time as $\hat{t}$, we write $t \leq \hat{t}$.

We define for each edge $(e_s, r, e_o, t)$ an inverse edge $(e_o, r^{-1}, e_s, t)$ that interchanges the positions of the subject and object entity to allow the random walker to move along the edge in both directions. The relation $r^{-1} \in \mathcal{R}$ is called the inverse relation of $r$.

\paragraph{Link forecasting}
The goal of the \textit{link forecasting} task is to predict new links for future timestamps. Given a query with a previously unseen timestamp $(e_s, r, ?, t)$, we want to identify a ranked list of object candidates that are most likely to complete the query. For subject prediction, we formulate the query as $(e_o, r^{-1}, ?, t)$.

\paragraph{Temporal random walk}
A \textit{non-increasing temporal random walk} $W$ of length $l \in \mathbb{N}$  from entity $e_{l+1} \in \mathcal{E}$ to entity $e_1 \in \mathcal{E}$ in the tKG $\mathcal{G}$ is defined as a sequence of edges
{\fontsize{9}{10}
\begin{align}
\begin{split}
((e_{l+1}, r_l, e_l, t_l)&, (e_l, r_{l-1}, e_{l-1}, t_{l-1}), \dots, (e_2, r_1, e_1, t_1))\\
& \text{with}\;t_l \geq t_{l-1} \geq \dots \geq t_1,
\label{eq:temporal_random_walk}
\end{split}
\end{align}}
where $(e_{i+1}, r_i, e_i, t_i) \in \mathcal{G}$ for $i \in [l]$. 

A non-increasing temporal random walk complies with time constraints so that the edges are traversed only backward in time, where it is also possible to walk along edges with the same timestamp. 

\paragraph{Temporal logical rule}
Let $E_i$ and $T_i$ for $i \in [l+1]$ be variables that represent entities and timestamps, respectively. Further, let $r_1, r_2, \dots, r_l, r_h \in \mathcal{R}$ be fixed.

A \textit{cyclic temporal logical rule} $R$ of length $l \in \mathbb{N}$ is defined as
{\fontsize{9}{10}
\begin{equation*}
((E_1, r_h, E_{l+1}, T_{l+1}) \leftarrow \wedge_{i=1}^l (E_i, r_i, E_{i+1}, T_i))
\label{eq:logical_rule}
\end{equation*}}
with the temporal constraints
{\fontsize{9}{10}
\begin{equation}
T_1 \leq T_2 \leq \dots \leq T_l < T_{l+1}.
\label{eq:rule_time_constraints}
\end{equation}}
The left-hand side of $R$ is called the rule head, with $r_h$ being the head relation, while the right-hand side is called the rule body, which is represented by a conjunction of body atoms $(E_i, r_i, E_{i+1}, T_i)$. The rule is called cyclic because the rule head and the rule body constitute two different walks connecting the same two variables $E_1$ and $E_{l+1}$.
A temporal rule implies that if the rule body holds with the temporal constraints given by~\eqref{eq:rule_time_constraints}, then the rule head is true as well for a future timestamp $T_{l+1}$.

The replacement of the variables $E_i$ and $T_i$ by constant terms is called grounding or instantiation.
For example, a grounding of the temporal rule
{\fontsize{9}{10}
$$
((E_1, \textit{consult}, E_2, T_2) \leftarrow (E_1, \textit{discuss by telephone}, E_2, T_1))
$$}
is given by the edges \textit{(Angela Merkel, discuss by telephone, Barack Obama, 2014/07/22)} and \textit{(Angela Merkel, consult, Barack Obama, 2014/08/09)} in Figure~\ref{fig:tkg_example}.
Let rule grounding refer to the replacement of the variables in the entire rule and body grounding refer to the replacement of the variables only in the body, where all groundings must comply with the temporal constraints in \eqref{eq:rule_time_constraints}. 

In many domains, logical rules are frequently violated so that confidence values are determined to estimate the probability of a rule's correctness. 
We adapt the standard confidence to take timestamp values into account. Let $(r_1, r_2, \dots, r_l, r_h)$ be the relations in a rule $R$. The body support is defined as the number of body groundings,
i.\,e., the number of tuples $(e_1, \dots, e_{l+1}, t_1, \dots, t_l)$ such that $(e_i, r_i, e_{i+1}, t_i) \in \mathcal{G}$ for $i \in [l]$ and $t_i \leq t_{i+1}$ for $i \in [l-1]$.
The rule support is defined as the number of body groundings such that there exists a timestamp $t_{l+1} > t_l$ with $(e_1, r_h, e_{l+1}, t_{l+1}) \in \mathcal{G}$.
The confidence of the rule $R$, denoted by conf($R$), can then be obtained by dividing the rule support by the body support.

\section{Our Framework}
We introduce TLogic, a rule-based link forecasting framework for tKGs. TLogic first extracts temporal walks from the graph and then lifts these walks to a more abstract, semantic level to obtain temporal rules that generalize to new data. 
The application of these rules generates answer candidates, for which the body groundings in the graph serve as explicit and human-readable explanations. 
Our framework consists of the components rule learning and rule application. The pseudocode for rule learning is shown in Algorithm~\ref{alg:rule_learning} and for rule application in Algorithm~\ref{alg:rule_application}.

\subsection{Rule Learning}
\begin{algorithm}[t]
\caption{Rule learning}
\label{alg:rule_learning}
\textbf{Input}: Temporal knowledge graph $\mathcal{G}$.\\
\textbf{Parameters}: Rule lengths $\mathcal{L} \subset \mathbb{N}$, number of temporal random walks $n \in \mathbb{N}$, transition distribution $d \in \{\mathrm{unif}, \exp\}$.\\
\textbf{Output}:  Temporal logical rules $\mathcal{TR}$.

\begin{algorithmic}[1] 
\FOR{relation $r \in \mathcal{R}$} 
\FOR{$l \in \mathcal{L}$}
\FOR{$i \in [n]$}
\STATE{$\mathcal{TR}_r^l \leftarrow \emptyset$}
\STATE {According to transition distribution $d$, sample a temporal random walk $W$ of length \mbox{$l+1$} with $t_{l+1} > t_l$. 
\hfill $\triangleright$~See~\eqref{eq:temporal_walk}.} 
\STATE{Transform walk $W$ to the corresponding temporal logical rule $R$. \hfill $\triangleright$~See~\eqref{eq:temporal_rule}.}
\STATE{Estimate the confidence of rule $R$. 

}
\STATE{$\mathcal{TR}_r^l \leftarrow \mathcal{TR}_r^l \cup \{(R, \text{conf}(R))\}$}
\ENDFOR
\ENDFOR
\STATE{$\mathcal{TR}_r \leftarrow \cup_{l \in \mathcal{L}} \mathcal{TR}_r^l$}
\ENDFOR
\STATE{$\mathcal{TR} \leftarrow \cup_{r \in \mathcal{R}} \mathcal{TR}_r$}
\STATE \textbf{return} $\mathcal{TR}$
\end{algorithmic}
\end{algorithm}
As the first step of rule learning, temporal walks are extracted from the tKG $\mathcal{G}$.
For a rule of length $l$, a walk of length $l+1$ is sampled, where the additional step corresponds to the rule head.

Let $r_h$ be a fixed relation, for which we want to learn rules. For the first sampling step $m=1$, we sample an edge $(e_1, r_h, e_{l+1}, t_{l+1})$, which will serve as the rule head, uniformly from all edges with relation type $r_h$. A temporal random walker then samples iteratively edges adjacent to the current object until a walk of length $l+1$ is obtained. 

For sampling step $m \in \{2, \dots, l+1\}$, let $(e_s, \tilde{r}, e_o, t)$ denote the previously sampled edge and $\mathcal{A}(m,e_o,t)$ the set of feasible edges for the next transition. To fulfill the temporal constraints in \eqref{eq:temporal_random_walk} and \eqref{eq:rule_time_constraints}, we define
{\fontsize{9}{10}
\begin{align*}
&\mathcal{A}(m, e_o, t) := \\
     &\begin{cases}
       \{(e_o, r, e, \hat{t}) \mid (e_o, r, e, \hat{t}) \in \mathcal{G},\; \hat{t} < t\} &\;\text{if}\; m = 2,\\
       \{(e_o, r, e, \hat{t}) \mid (e_o, r, e, \hat{t}) \in \tilde{\mathcal{G}},\; \hat{t} \leq t\} &\;\text{if}\; m \in \{3, \dots, l\},\\
       \{(e_o, r, e_1, \hat{t}) \mid (e_o, r, e_1, \hat{t}) \in \tilde{\mathcal{G}},\; \hat{t} \leq t\} &\;\text{if}\; m = l+1,
     \end{cases}
\end{align*}}
where $\tilde{\mathcal{G}} := \mathcal{G}\setminus \{(e_o, \tilde{r}^{-1}, e_s, t)\}$ excludes the inverse edge to avoid redundant rules. For obtaining cyclic walks, we sample in the last step $m = l+1$ an edge that connects the walk to the first entity $e_1$ if such edges exist. 
Otherwise, we sample the next walk.

The transition distribution for sampling the next edge can either be uniform or exponentially weighted. 
We define an index mapping $\hat{m} := (l+1) - (m-2)$ to be consistent with the indices in \eqref{eq:temporal_random_walk}. Then, the exponentially weighted probability for choosing edge $u \in \mathcal{A}\left(m,e_{\hat{m}}, t_{\hat{m}}\right)$ for $m \in \{2, \dots, l+1\}$ is given by 
{
\begin{equation}
\mathbb{P}(u; m, e_{\hat{m}}, t_{\hat{m}}) = \frac{\exp(t_u - t_{\hat{m}})}{\sum\limits_{\hat{u} \in \mathcal{A}\left(m, e_{\hat{m}}, t_{\hat{m}}\right)}\exp(t_{\hat{u}} - t_{\hat{m}})}
\label{eq:exp_distribution}
\end{equation}}
where $t_u$ denotes the timestamp of edge $u$.
The exponential weighting favors edges with timestamps that are closer to the timestamp of the previous edge and probably more relevant for prediction. 

The resulting temporal walk $W$ is given by
{\fontsize{9}{10}
\begin{equation}
((e_1, r_h, e_{l+1}, t_{l+1}), (e_{l+1}, r_l, e_l, t_l), \dots, (e_2, r_1, e_1, t_1)).
\label{eq:temporal_walk}
\end{equation}}
$W$ can then be transformed to a temporal rule $R$ by replacing the entities and timestamps with variables. While the first edge in $W$ becomes the rule head $(E_1, r_h, E_{l+1}, T_{l+1})$, the other edges are mapped to body atoms, where each edge $(e_{i+1}, r_i, e_i, t_i)$ is converted to the body atom $(E_i, r_i^{-1}, E_{i+1}, T_i)$. The final rule $R$ is denoted by
{\fontsize{9}{10}
\begin{equation}
((E_1, r_h, E_{l+1}, T_{l+1}) \leftarrow \wedge_{i=1}^l (E_i, r_i^{-1}, E_{i+1}, T_i)).
\label{eq:temporal_rule}
\end{equation}}
In addition, we impose the temporal consistency constraints  $T_1 \leq T_2 \leq \dots \leq T_l < T_{l+1}$. 

The entities $(e_1, \dots, e_{l+1})$ in $W$ do not need to be distinct since a pair of entities can have many interactions at different points in time. For example, Angela Merkel made several visits to China in 2014, which could constitute important information for the prediction. Repetitive occurrences of the same entity in $W$ are replaced with the same random variable in $R$ to maintain this knowledge.

For the confidence estimation of $R$, we sample from the graph a fixed number of body groundings, which have to match the body relations and the variable constraints mentioned in the last paragraph while satisfying the condition from~\eqref{eq:rule_time_constraints}.
The number of unique bodies serves as the body support. The rule support is determined by counting the number of bodies for which an edge with relation type $r_h$ exists that connects $e_1$ and $e_{l+1}$ from the body. Moreover, the timestamp of this edge has to be greater than all body timestamps to fulfill~\eqref{eq:rule_time_constraints}. 

For every relation $r \in \mathcal{R}$, we sample $n \in \mathbb{N}$ temporal walks for a set of prespecified lengths $\mathcal{L} \subset \mathbb{N}$. The set $\mathcal{TR}_r^l$ stands for all rules of length $l$ with head relation $r$ with their corresponding confidences. All rules for relation $r$ are included in $\mathcal{TR}_r := \cup_{l \in \mathcal{L}} \mathcal{TR}_r^l$, and the complete set of learned temporal rules is given by $\mathcal{TR} := \cup_{r \in \mathcal{R}} \mathcal{TR}_r$. 

It is possible to learn rules only for a single relation or a set of specific relations of interest.
Explicitly learning rules for all relations is especially effective for rare relations that would otherwise only be sampled with a small probability.
The learned rules are not specific to the graph from which they have been extracted, but they could be employed in an inductive setting where the rules are transferred to related datasets that share a common vocabulary for straightforward application.

\subsection{Rule Application}
\begin{algorithm}[t]
\caption{Rule application}
\label{alg:rule_application}
\textbf{Input}: Test query $q = (e^q, r^q, ?, t^q)$, temporal logical rules $\mathcal{TR}$, temporal knowledge graph $\mathcal{G}$.\\
\textbf{Parameters}: Time window $w \in \mathbb{N} \cup \{\infty\}$, minimum number of candidates $k$, score function $f$.\\
\textbf{Output}: Answer candidates $\mathcal{C}$.

\begin{algorithmic}[1] 
\STATE{$\mathcal{C} \leftarrow \emptyset$

$\triangleright$ Apply the rules in $\mathcal{TR}$ by decreasing confidence.}
\IF{$\mathcal{TR}_{r^q} \neq \emptyset$ }
\FOR{rule $R \in \mathcal{TR}_{r^q}$}
\STATE{Find all body groundings of $R$ in $\mathcal{SG} \subset \mathcal{G}$, where $\mathcal{SG}$ consists of the edges within the time window $[t^q-w, t^q)$.} 
\STATE{Retrieve candidates $\mathcal{C}(R)$ from the target entities of the body groundings.}
\FOR{$c \in \mathcal{C}(R)$}
\STATE{Calculate score $f(R,c)$. \hfill $\triangleright$ See~\eqref{eq:score_function}.}
\STATE{$\mathcal{C} \leftarrow \mathcal{C} \cup \{(c, f(R,c))\}$}
\ENDFOR
\IF{$|\{c \mid \exists R: (c, f(R,c)) \in \mathcal{C}\}| \geq k$}
\STATE{\textbf{break}}
\ENDIF
\ENDFOR
\ENDIF
\STATE \textbf{return} $\mathcal{C}$
\end{algorithmic}
\end{algorithm}
The learned temporal rules $\mathcal{TR}$ are applied to answer queries of the form $q = (e^q, r^q, ?, t^q)$. The answer candidates are retrieved from the target entities of body groundings in the tKG $\mathcal{G}$. 
If there exist no rules $\mathcal{TR}_{r^q}$ for the query relation $r^q$, or if there are no matching body groundings in the graph, then no answers are predicted for the given query.

To apply the rules on relevant data, a subgraph $\mathcal{SG} \subset \mathcal{G}$ dependent on a time window $w \in \mathbb{N} \cup \{\infty\}$ is retrieved. For $w \in \mathbb{N}$, the subgraph $\mathcal{SG}$ contains all edges from $\mathcal{G}$ that have timestamps $t \in [t^q-w, t^q)$. If $w = \infty$, then all edges with timestamps prior to the query timestamp $t^q$ are used for rule application, i.\,e., $\mathcal{SG}$ consists of all facts with $t \in [t_{\mathrm{min}}, t^q)$, where $t_{\mathrm{min}}$ is the minimum timestamp in the graph $\mathcal{G}$.

We apply the rules $\mathcal{TR}_{r^q}$ by decreasing confidence, where each rule $R$ generates a set of answer candidates $\mathcal{C}(R)$.
Each candidate $c \in \mathcal{C}(R)$ is then scored by a function $f: \mathcal{TR}_{r^q} \times \mathcal{E} \rightarrow [0,1]$ that reflects the probability of the candidate being the correct answer to the query.

Let $\mathcal{B}(R,c)$ be the set of body groundings of rule $R$ that start at entity $e^q$ and end at entity $c$. 
We choose as score function $f$ a convex combination of the rule's confidence and a function that takes the time difference $t^q - t_1(\mathcal{B}(R,c))$ as input, where  $t_1(\mathcal{B}(R,c))$ denotes the earliest timestamp $t_1$ in the body. If several body groundings exist, we take from all possible $t_1$ values the one that is closest to $t^q$.
For candidate $c \in \mathcal{C}(R)$, the score function is defined as
{\fontsize{9}{10}
\begin{equation}
f(R,c) = a \cdot \mathrm{conf}(R) + (1-a) \cdot \exp(-\lambda (t^q - t_1(\mathcal{B}(R,c))))
\label{eq:score_function}
\end{equation}}
with $\lambda > 0$ and $a \in [0,1]$. 

The intuition for this choice of $f$ is that candidates generated by high-confidence rules should receive a higher score. Adding a dependency on the timeframe of the rule grounding is based on the observation that the existence of edges in a rule become increasingly probable with decreasing time difference between the edges.
We choose the exponential distribution since it is commonly used to model interarrival times of events. The time difference $t^q - t_1(\mathcal{B}(R,c))$ is always non-negative for a future timestamp value $t^q$, and with the assumption that there exists a fixed mean, the exponential distribution is also the maximum entropy distribution for such a time difference variable. The exponential distribution is rescaled so that both summands are in the range $[0,1]$.

All candidates are saved with their scores as $(c, f(R,c))$ in $\mathcal{C}$.
We stop the rule application when the number of different answer candidates $|\{c \mid \exists R: (c, f(R,c)) \in \mathcal{C}\}|$ is at least $k$ so that there is no need to go through all rules. 

\subsection{Candidate Ranking}
For the ranking of the answer candidates, all scores of each candidate $c$ are aggregated through a noisy-OR calculation, which produces the final score
{
\begin{equation}
1 - \Pi_{\{s \mid (c, s) \in \mathcal{C}\}} (1 - s).
\label{eq:score_aggregation}
\end{equation}}
The idea is to aggregate the scores to produce a probability, where candidates implied by more rules should have a higher score.

In case there are no rules for the query relation $r^q$, or if there are no matching body groundings in the graph, it might still be interesting to retrieve possible answer candidates. 
In the experiments, we apply a simple baseline where the scores for the candidates are obtained from the overall object distribution in the training data if $r^q$ is a new relation. If $r^q$ already exists in the training set, we take the object distribution of the edges with relation type $r^q$. 

\section{Experiments}

\subsection{Datasets}
We conduct experiments on the dataset Integrated Crisis Early Warning System\footnote{\url{https://dataverse.harvard.edu/dataverse/icews}} (ICEWS), which contains information about international events and is a commonly used benchmark dataset for link prediction on tKGs. 
We choose the subsets ICEWS14, ICEWS18, and ICEWS0515, which include data from the years 2014, 2018, and 2005 to 2015, respectively. 
Since we consider link forecasting, each dataset is split into training, validation, and test set so that the timestamps in the training set occur earlier than the timestamps in the validation set, which again occur earlier than the timestamps in the test set. 
To ensure a fair comparison, we use the split provided by~\citet{xerte.han.2021}\footnote{\url{https://github.com/TemporalKGTeam/xERTE}}.
The statistics of the datasets are summarized in the supplementary material.

\subsection{Experimental Setup}
For each test instance $(e_s^q, r^q, e_o^q, t^q)$, we generate a list of candidates for both object prediction $(e_s^q, r^q, ?, t^q)$ and subject prediction $(e_o^q, (r^q)^{-1}, ?, t^q)$. The candidates are ranked by decreasing scores, which are calculated according to \eqref{eq:score_aggregation}.

The confidence for each rule is estimated by sampling $500$ body groundings and counting the number of times the rule head holds. We learn rules of the lengths 1, 2, and 3, and for application, we only consider the rules with a minimum confidence of $0.01$ and minimum body support of $2$.

We compute the mean reciprocal rank (MRR) and hits@$k$ for $k \in \{1, 3, 10\}$, which are standard metrics for link prediction on KGs. For a rank $x \in \mathbb{N}$, the reciprocal rank is defined as $\frac{1}{x}$, and the MRR is the average of all reciprocal ranks of the correct query answers across all queries.  
The metric hits@$k$ (h@$k$) indicates the proportion of queries for which the correct entity appears under the top $k$ candidates.

Similar to~\citet{xerte.han.2021}, we perform time-aware filtering where all correct entities at the query timestamp except for the true query object are filtered out from the answers. In comparison to the alternative setting that filters out all other objects that appear together with the query subject and relation at any timestamp, time-aware filtering yields a more realistic performance estimate. 

\paragraph{Baseline methods}
We compare TLogic\footnote{Code available at \url{https://github.com/liu-yushan/TLogic.}} with the state-of-the-art baselines for static link prediction
DistMult~\cite{distmult.yang.2015}, ComplEx~\cite{complex.trouillon.2016}, and AnyBURL~\cite{anyburl.meilicke.2019, anyburl.meilicke.2020} as well as for temporal link prediction 
TTransE~\cite{ttranse.leblay.2018}, TA-DistMult~\cite{ta-distmult-transe.garcia-duran.2018}, DE-SimplE~\cite{de-simple.goel.2020}, TNTComplEx~\cite{tntcomplex.lacroix.2020}, CyGNet~\cite{cygnet.zhu.2021}, RE-Net~\cite{renet.jin.2019}, and xERTE~\cite{xerte.han.2021}. 
All baseline results except for the results on AnyBURL are from~\citet{xerte.han.2021}. AnyBURL samples paths based on reinforcement learning and generalizes them to rules, where the rule space also includes, e.\,g., acyclic rules and rules with constants. A non-temporal variant of TLogic would sample paths randomly and only learn cyclic rules, which would presumably yield worse performance than AnyBURL. Therefore, we choose AnyBURL as a baseline to assess the effectiveness of adding temporal constraints. 
\subsection{Results}
\begin{table*}[t]
\centering
\small
\begin{tabular}{l|cccc|cccc|cccc}
\toprule
Dataset & \multicolumn{4}{|c}{\textbf{ICEWS14}} &  \multicolumn{4}{|c}{\textbf{ICEWS18}} & \multicolumn{4}{|c}{\textbf{ICEWS0515}}\\
 \midrule
Model & MRR & h@1 & h@3 & h@10 & MRR & h@1 & h@3 & h@10 & MRR & h@1 & h@3 & h@10 \\
\midrule
DistMult & 0.2767 & 0.1816 & 0.3115 & 0.4696 & 0.1017 & 0.0452 & 0.1033 & 0.2125 & 0.2873 &  0.1933 &  0.3219 & 0.4754\\
ComplEx & 0.3084 & 0.2151 & 0.3448 & 0.4958 & 0.2101 & 0.1187 & 0.2347 & 0.3987 & 0.3169 & 0.2144 & 0.3574 & 0.5204\\
AnyBURL & 0.2967 & 0.2126 & 0.3333 & 0.4673 & 0.2277 & 0.1510 & 0.2544 & 0.3891 & 0.3205 & 0.2372 & 0.3545 & 0.5046\\
\midrule
TTransE & 0.1343 & 0.0311 & 0.1732 & 0.3455 & 0.0831 & 0.0192 & 0.0856 & 0.2189 & 0.1571 & 0.0500 & 0.1972 & 0.3802\\
TA-DistMult & 0.2647 & 0.1709 & 0.3022 & 0.4541 & 0.1675 & 0.0861 & 0.1841 & 0.3359 & 0.2431 & 0.1458 & 0.2792 & 0.4421\\
DE-SimplE & 0.3267 & 0.2443 & 0.3569 & 0.4911 & 0.1930 & 0.1153 & 0.2186 & 0.3480 & 0.3502 & 0.2591 & 0.3899 & 0.5275\\
TNTComplEx & 0.3212 & 0.2335 & 0.3603 & 0.4913 & 0.2123 & 0.1328 & 0.2402 & 0.3691 & 0.2754 & 0.1952 & 0.3080 & 0.4286\\
CyGNet & 0.3273 & 0.2369 & 0.3631 & 0.5067 & 0.2493 & 0.1590 & 0.2828 & 0.4261 & 0.3497 & 0.2567 & 0.3909 & 0.5294\\
RE-Net & 0.3828 & 0.2868 & 0.4134 & 0.5452 & 0.2881 & 0.1905 & 0.3244 & 0.4751 & 0.4297 & 0.3126 & 0.4685 & 0.6347\\
xERTE & 0.4079  & 0.3270 & 0.4567 & 0.5730 & 0.2931 & \textbf{0.2103} & 0.3351 & 0.4648 & 0.4662 & \textbf{0.3784} & 0.5231 & 0.6392\\
\midrule
TLogic & \textbf{0.4304} & \textbf{0.3356} & \textbf{0.4827} & \textbf{0.6123} & \textbf{0.2982} & 0.2054 & \textbf{0.3395} & \textbf{0.4853} & \textbf{0.4697} & 0.3621 & \textbf{0.5313} & \textbf{0.6743}\\
\bottomrule
\end{tabular}
\caption{Results of link forecasting on the datasets ICEWS14, ICEWS18, and ICEWS0515. All metrics are time-aware filtered. The best results among all models are displayed in bold.}
\label{tab:results}
\end{table*}
The results of the experiments are displayed in Table~\ref{tab:results}. TLogic outperforms all baseline methods with respect to the metrics MRR, hits@3, and hits@10. Only xERTE performs better than Tlogic for hits@1 on the datasets ICEWS18 and ICEWS0515. 

Besides a list of possible answer candidates with corresponding scores, TLogic can also provide temporal rules and body groundings in form of walks from the graph that support the predictions. 
Table~\ref{tab:ex_rules} presents three exemplary rules with high confidences that were learned from ICEWS14.
For the query \textit{(Angela Merkel, consult, ?, 2014/08/09)}, two walks are shown in Table~\ref{tab:ex_rules}, which serve as time-consistent explanations for the correct answer \textit{Barack Obama}.
\begin{table*}[t]
\centering
\small
\begin{tabular}{c|c|c}
\toprule
Confidence & Head & Body\\
\midrule
0.963 & $(E_1, \textit{demonstrate or rally}, E_2, T_4)$ &  $(E_1, \textit{riot}, E_2, T_1)$ $\wedge$  $(E_2, \textit{make statement}, E_1, T_2)$ $\wedge$  $(E_1, \textit{riot}, E_2, T_3)$\\
\midrule
0.818 & $(E_1, \textit{share information}, E_2, T_2)$ &  $(E_1, \textit{express intent to ease sanctions}^{-1}, E_2, T_1)$\\
\midrule
0.750 & $(E_1, \textit{provide military aid}, E_3, T_3)$ &  $(E_1, \textit{provide military aid}, E_2, T_1)$ $\wedge$  $(E_2, \textit{intend to protect}^{-1}, E_3, T_2)$\\
\midrule
0.570 & \textit{(Merkel, consult, Obama, 14/08/09)} & \textit{(Merkel, discuss by telephone, Obama, 14/07/22)}\\
\midrule
0.500 & \textit{(Merkel, consult, Obama, 14/08/09)} & \textit{(Merkel, express intent to meet,  Obama, 14/05/02)}\\ & & $\wedge$ $(\textit{Obama}, \textit{consult}^{-1}$, \textit{Merkel, 14/07/18)} $\wedge$ $(\textit{Merkel}, \textit{consult}^{-1}$, \textit{Obama, 14/07/29)}\\
\bottomrule
\end{tabular}
\caption{Three exemplary rules from the dataset ICEWS14 and two walks for the query \textit{(Angela Merkel, consult, ?, 2014/08/09)} that lead to the correct answer \textit{Barack Obama}. The timestamps are displayed in the format yy/mm/dd.}
\label{tab:ex_rules}
\end{table*}
\begin{figure}[t]
\centering
\includegraphics[width=0.8\columnwidth]{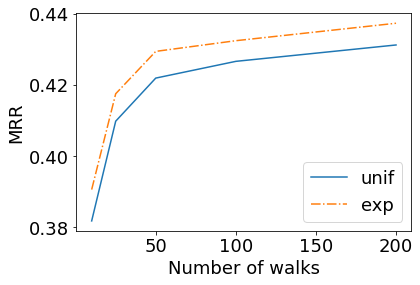}
\caption{MRR performance on the validation set of ICEWS14. The transition distribution is either uniform or exponentially weighted.}
\label{fig:mrr}
\end{figure}

\paragraph{Inductive setting}
One advantage of our learned logical rules is that they are applicable to any new dataset as long as the new dataset covers common relations. This might be relevant for cases where new entities appear. For example, Donald Trump, who served as president of the United States from 2017 to 2021, is included in the dataset ICEWS18 but not in ICEWS14. The logical rules are not tied to particular entities and would still be applicable, while embedding-based methods have difficulties operating in this challenging setting. The models would need to be retrained to obtain embeddings for the new entities, where existing embeddings might also need to be adapted to the different time range.

For the two rule-based methods AnyBURL and TLogic, we apply the rules learned on the training set of ICEWS0515 (with timestamps from 2005/01/01 to 2012/08/06) to the test set of ICEWS14 as well as the rules learned on the training set of ICEWS14 to the test set of ICEWS18 (see Table~\ref{tab:inductive_setting}).
The performance of TLogic in the inductive setting is for all metrics close to the results in Table~\ref{tab:results}, while for AnyBURL, especially the results on ICEWS18 drop significantly. It seems that the encoded temporal information in TLogic is essential for achieving correct predictions in the inductive setting.
ICEWS14 has only 7,128 entities, while ICEWS18 contains 23,033 entities. The results confirm that temporal rules from TLogic can even be transferred to a dataset with a large number of new entities and timestamps and lead to a strong performance.

\subsection{Analysis}
The results in this section are obtained on the dataset ICEWS14, but the findings are similar for the other two datasets.
More detailed results can be found in the supplementary material.
\begin{table*}[t]
\centering
\small
\begin{tabular}{c c|c|c c c c c}
\toprule
$\mathcal{G}_{\mathrm{train}}$ & $\mathcal{G}_{\mathrm{test}}$ & Model & MRR & h@1 & h@3 & h@10\\
\midrule
\multirow{2}{*}{ICEWS0515} & \multirow{2}{*}{ICEWS14} & AnyBURL & 0.2664 & 0.1800 & 0.3024 & 0.4477\\
& & TLogic & 0.4253 & 0.3291 & 0.4780 & 0.6122\\ 
\midrule  
\multirow{2}{*}{ICEWS14} & \multirow{2}{*}{ICEWS18} & AnyBURL & 0.1546 & 0.0907 & 0.1685 & 0.2958\\
& & TLogic & 0.2915 & 0.1987 & 0.3330 & 0.4795\\ 
\bottomrule
\end{tabular}
\caption{Inductive setting where rules learned on $\mathcal{G}_{\mathrm{train}}$ are transferred and applied to $\mathcal{G}_{\mathrm{test}}$.}
\label{tab:inductive_setting}
\end{table*}

\paragraph{Number of walks}
Figure~\ref{fig:mrr} shows the MRR performance on the validation set of ICEWS14 for different numbers of walks that were extracted during rule learning. We observe a performance increase with a growing number of walks. However, the performance gains saturate between 100 and 200 walks where rather small improvements are attainable.

\paragraph{Transition distribution}
We test two transition distributions for the extraction of temporal walks: uniform and exponentially weighted according to~\eqref{eq:exp_distribution}. The rationale behind using an exponentially weighted distribution is the observation that related events tend to happen within a short timeframe. 
The distribution of the first edge is always uniform to not restrict the variety of obtained walks.
Overall, the performance of the exponential distribution consistently exceeds the uniform setting with respect to the MRR (see Figure~\ref{fig:mrr}).

We observe that the exponential distribution leads to more rules of length 3 than the uniform setting (11,718 compared to 8,550 rules for 200 walks), while it is the opposite for rules of length 1 (7,858 compared to 11,019 rules). The exponential setting leads to more successful longer walks because the timestamp differences between subsequent edges tend to be smaller. It is less likely that there are no feasible transitions anymore because of temporal constraints. The uniform setting, however, leads to a better exploration of the neighborhood around the start node for shorter walks.

\paragraph{Rule length}
We learn rules of lengths 1, 2, and 3. Using all rules for application results in the best performance (MRR on the validation set: 0.4373), followed by rules of only length 1 (0.4116), 3 (0.4097), and 2 (0.1563). The reason why rules of length 3 perform better than length 2 is that the temporal walks are allowed to transition back and forth between the same entities. Since we only learn cyclic rules, a rule body of length 2 must constitute a path with no recurring entities, resulting in fewer rules and rule groundings in the graph. Interestingly, simple rules of length 1 already yield very good performance.   

\paragraph{Time window}
For rule application, we define a time window for retrieving the relevant data. The performance increases with the size of the time window, even though relevant events tend to be close to the query timestamp. The second summand of the score function $f$ in \eqref{eq:score_function} takes the time difference between the query timestamp $t^q$ and the earliest body timestamp $t_1(\mathcal{B}(R,c))$ into account. In this case, earlier events with a large timestamp difference receive a lesser weight, while generally, as much information as possible is beneficial for prediction.

\paragraph{Score function}
We define the score function $f$ in \eqref{eq:score_function} as a convex combination of the rule's confidence and a function that depends on the time difference $t^q - t_1(\mathcal{B}(R,c))$. 
The performance of only using the confidence (MRR: 0.3869) or only using the exponential function (0.4077) is worse than the combination (0.4373), which means that both the information from the rules' confidences and the time differences are important for prediction.





\paragraph{Variance}
The variance in the performance due to different rules obtained from the rule learning component is quite small. 
Running the same model with the best hyperparameter settings for five different seeds results in a standard deviation of 0.0012 for the MRR.
The rule application component is deterministic and always leads to the same candidates with corresponding scores for the same hyperparameter setting.

\paragraph{Training and inference time}
The worst-case time complexity for learning rules of length $l$ is $\mathcal{O}(|\mathcal{R}|nlDb)$, where $n$ is the number of walks, $D$ the maximum node degree in the training set, and $b$ the number of body samples for estimating the confidence. 
The worst-case time complexity for inference is given by $\mathcal{O}(|\mathcal{G}| + |\mathcal{TR}_{r^q}|D^L|\mathcal{E}|\log(k))$, where $L$ is the maximum rule length in $\mathcal{TR}_{r^q}$ and $k$ the minimum number of candidates. For large graphs with high node degrees, it is possible to reduce the complexity to $ \mathcal{O}\left(|\mathcal{G}| +\vert\mathcal{TR}_{r^q}\vert KLD|\mathcal{E}| \log(k) \right)$ by only keeping a maximum of $K$ candidate walks during rule application.

Both training and application can be parallelized since the rule learning for each relation and the rule application for each test query are independent.
Rule learning with 200 walks and exponentially weighted transition distribution for rule lengths $\{1,2,3\}$ on a machine with 8 CPUs takes 180 sec for ICEWS14, while the application on the validation set takes 2000 sec, with $w=\infty$ and $k = 20$. For comparison, the best-performing baseline xERTE needs for training one epoch on the same machine already 5000 sec, where an MRR of 0.3953 can be obtained, while testing on the validation set takes 700 sec.  


\section{Conclusion}
We have proposed TLogic, the first symbolic framework that directly learns temporal logical rules from temporal knowledge graphs and applies these rules for link forecasting.
The framework generates answers by applying rules to observed events prior to the query timestamp and scores the answer candidates depending on the rules' confidences and time differences. 
Experiments on three datasets indicate that TLogic achieves better overall performance compared to state-of-the-art baselines.
In addition, our approach also provides time-consistent, explicit, and human-readable explanations for the predictions in the form of temporal logical rules.

As future work, it would be interesting to
integrate acyclic rules, which could also contain relevant information and might boost the performance for rules of length 2.
Furthermore, the simple sampling mechanism for temporal walks could be replaced by a more sophisticated approach, which is able to effectively identify the most promising walks.
 
\section*{Acknowledgement}
This work has been supported by the German Federal Ministry for Economic Affairs and Climate Action (BMWK) as part of the project RAKI under grant number 01MD19012C and by the German Federal Ministry of Education and Research (BMBF) under grant number 01IS18036A. The authors of this work take full responsibility for its content.
 
\bibliography{aaai22.bib}

\newpage
\appendix

\section{Supplementary Material}

\paragraph{Dataset statistics}
Table \ref{tab:dataset_statistics} shows the statistics of the three datasets ICEWS14, ICEWS18, and ICEWS0515. $|\mathcal{X}|$ denotes the cardinality of a set $\mathcal{X}$.

\begin{table}[h]
\centering
\small
\resizebox{\columnwidth}{!}{
\begin{tabular}{c|c c c c c c}
\toprule
Dataset & $|\mathcal{G}_{\mathrm{train}}|$ & $|\mathcal{G}_{\mathrm{valid}}|$ & $|\mathcal{G}_{\mathrm{test}}|$ & $|\mathcal{E}|$ & $|\mathcal{R}|$ & $|\mathcal{T}|$\\
\midrule
14 & 63,685 & 13,823 & 13,222 & 7,128 & 230 & 365\\
18 & 373,018 & 45,995 & 49,545 & 23,033 & 256 & 304\\
0515 & 322,958 & 69,224 & 69,147 & 10,488 & 251 & 4,017\\
\bottomrule
\end{tabular}}
\caption{Dataset statistics with daily time resolution for all three ICEWS datasets.}
\label{tab:dataset_statistics}
\end{table}

\paragraph{Experimental details}
All experiments were conducted on a Linux machine with 16 CPU cores and 32 GB RAM.
The set of tested hyperparameter ranges and best parameter values for TLogic are displayed in Table \ref{tab:hyperparameters}. Due to memory constraints, the time window $w$ for ICEWS18 is set to 200 and for ICEWS0515 to 1000.
The best hyperparameter values are chosen based on the MRR on the validation set.
Due to the small variance of our approach, the shown results are based on one algorithm run. A random seed of 12 is fixed for the rule learning component to obtain reproducible results.

\begin{table}[h]
\centering
\small
\resizebox{\columnwidth}{!}{
\begin{tabular}{c|c c}
\toprule
Hyperparameter & Range & Best\\
\midrule
Number of walks $n$ & $\{10, 25, 50, 100, 200\}$ & 200\\
\midrule
Transition distribution $d$ & $\{\text{unif}, \exp\}$ & $\exp$\\
\midrule
Rule lengths $\mathcal{L}$ & $\{\{1\}, \{2\}, \{3\}, \{1,2,3\}\}$ & $\{1,2,3\}$\\
\midrule
Time window $w$ & $\{30, 90, 150, 210, 270, \infty\}$ & $\infty$\\
\midrule
Minimum candidates $k$ & $\{10, 20\}$ & 20\\
\midrule
$\alpha$ (score function $f$) & $\{0, 0.25, 0.5, 0.75, 1\}$ & 0.5\\
\midrule
$\lambda$ (score function $f$) & $\{0.01, 0.1, 0.5, 1\}$ & 0.1\\
\bottomrule
\end{tabular}}
\caption{Hyperparameter ranges and best parameter values.}
\label{tab:hyperparameters}
\end{table}

All results in the appendix refer to the validation set of ICEWS14. However, the observations are similar for the test set and the other two datasets. All experiments use the best set of hyperparameters, where only the analyzed parameters are modified. 

\paragraph{Object distribution baseline}
We apply a simple object distribution baseline when there are no rules for the query relation or no matching body groundings in the graph. This baseline is only added for completeness and does not improve the results in a significant way. 

The proportion of cases where there are no rules for the test query relation is 15/26,444 = 0.00056 for ICEWS14, 21/99,090 = 0.00021 for ICEWS18, and 9/138,294 = 0.00007 for ICEWS0515.
The proportion of cases where there are no matching body groundings is 880/26,444 = 0.0333 for ICEWS14, 2,535/99,090 = 0.0256 for ICEWS18, and 2,375/138,294 = 0.0172 for ICEWS0515.

\paragraph{Number of walks and transition distribution}
Table \ref{tab:num_walks_transition} shows the results for different choices of numbers of walks and transition distributions.
The performance for all metrics increases with the number of walks. Exponentially weighted transition always outperforms uniform sampling.
\begin{table}[h]
\centering
\resizebox{\columnwidth}{!}{
\begin{tabular}{c c|c c c c c}
\toprule
Walks & Transition & MRR & h@1 & h@3 & h@10\\
\midrule
10 & Unif & 0.3818 & 0.2983 & 0.4307 & 0.5404\\
10 & Exp & 0.3906 & 0.3054 & 0.4408 & 0.5530\\
\midrule \relax 
25 & Unif & 0.4098 & 0.3196 & 0.4614 & 0.5803\\
25 & Exp & 0.4175 & 0.3270 & 0.4710 & 0.5875\\
\midrule \relax 
50 & Unif & 0.4219 & 0.3307 & 0.4754 & 0.5947\\
50 & Exp & 0.4294 & 0.3375 & 0.4837 & 0.6024\\
\midrule \relax 
100 & Unif & 0.4266 & 0.3315 & 0.4817 & 0.6057\\
100 & Exp & 0.4324 & 0.3397 & 0.4861 & 0.6092\\
\midrule \relax 
200 & Unif & 0.4312 & 0.3366 & 0.4851 & 0.6114\\
200 & Exp & 0.4373 & 0.3434 & 0.4916 & 0.6161\\
\bottomrule
\end{tabular}}
\caption{Results for different choices of numbers of walks and transition distributions.}
\label{tab:num_walks_transition}
\end{table}

\paragraph{Rule length}
Table \ref{tab:rule_length} indicates that using rules of all lengths for application results in the best performance. Learning only cyclic rules probably makes it more difficult to find rules of length 2, where the rule body must constitute a path with no recurring entities, leading to fewer rules and body groundings in the graph.
\begin{table}[h]
\centering
\resizebox{0.9\columnwidth}{!}{
\begin{tabular}{c|c c c c c}
\toprule
Rule length & MRR & h@1 & h@3 & h@10\\
\midrule
1 & 0.4116 & 0.3168 & 0.4708 & 0.5909\\
\midrule
2 & 0.1563 & 0.0648 & 0.1776 & 0.3597\\
\midrule
3 & 0.4097 & 0.3213 & 0.4594 & 0.5778\\
\midrule
1,2,3 & 0.4373 & 0.3434 & 0.4916 & 0.6161\\
\bottomrule
\end{tabular}}
\caption{Results for different choices of rule lengths.}
\label{tab:rule_length}
\end{table}

\paragraph{Time window}
Generally, the larger the time window, the better the performance (see Table \ref{tab:time_window}). If taking all previous timestamps leads to a too high memory usage, the time window should be decreased.
\begin{table}[h]
\centering
\resizebox{0.9\columnwidth}{!}{
\begin{tabular}{c|c c c c c}
\toprule
Time window & MRR & h@1 & h@3 & h@10\\
\midrule
30 & 0.3842 & 0.3080 & 0.4294 & 0.5281\\
\midrule
90 & 0.4137 & 0.3287 & 0.4627 & 0.5750\\
\midrule
150 & 0.4254 & 0.3368 & 0.4766 & 0.5950\\
\midrule
210 & 0.4311 & 0.3403 & 0.4835 & 0.6035\\
\midrule
270 & 0.4356 & 0.3426 & 0.4892 & 0.6131\\
\midrule
$\infty$ & 0.4373 & 0.3434 & 0.4916 & 0.6161\\
\bottomrule
\end{tabular}}
\caption{Results for different choices of time windows.}
\label{tab:time_window}
\end{table}

\paragraph{Score function}
Using the best hyperparameters values for $\alpha$ and $\lambda$, Table \ref{tab:score_function} shows in the first row the results if only the rules' confidences are used for scoring, in the second row if only the exponential component is used, and in the last row the results for the combined score function. The combination yields the best overall performance. The optimal balance between the two terms, however, depends on the application and metric prioritization.
\begin{table}[h]
\centering
\resizebox{0.9\columnwidth}{!}{
\begin{tabular}{c c|c c c c}
\toprule
$\alpha$ & $\lambda$ & MRR & h@1 & h@3 & h@10\\
\midrule
$1$ & arbitrary & 0.3869 & 0.2806 & 0.4444 & 0.5918\\
\midrule
$0$ & 0.1 & 0.4077 & 0.3515 & 0.4820 & 0.6051\\
\midrule
$0.5$ & 0.1 & 0.4373 & 0.3434 & 0.4916 & 0.6161\\
\bottomrule
\end{tabular}}
\caption{Results for different parameter values in the score function $f$.}
\label{tab:score_function}
\end{table}

\begin{figure}[!h]
\centering
\includegraphics[width=0.8\columnwidth]{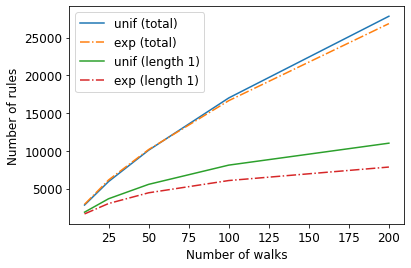}
\caption{Total number of learned rules and number of rules for length 1.}
\label{fig:number_rules_1}
\end{figure}

\begin{figure}[!h]
\centering
\includegraphics[width=0.8\columnwidth]{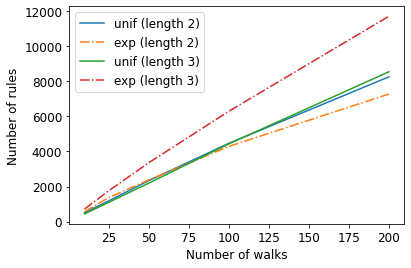}
\caption{Number of rules for lengths 2 and 3.}
\label{fig:number_rules_2}
\end{figure}

\paragraph{Rule learning}
The figures \ref{fig:number_rules_1} and \ref{fig:number_rules_2} show the number of rules learned under the two transition distributions. The total number of learned rules is similar for the uniform and exponential distribution, but there is a large difference for rules of lengths 1 and 3. 
The exponential distribution leads to more successful longer walks and thus more longer rules, while the uniform distribution leads to a better exploration of the neighborhood around the start node for shorter walks.

\begin{figure}[!h]
\centering
\includegraphics[width=0.8\columnwidth]{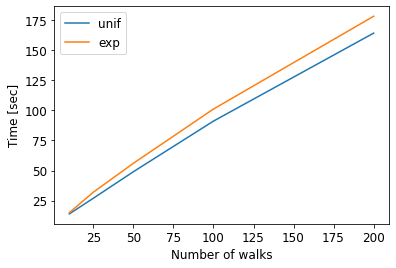}
\caption{Rule learning time.}
\label{fig:learning_time}
\end{figure}

\begin{figure}[!h]
\centering
\includegraphics[width=0.8\columnwidth]{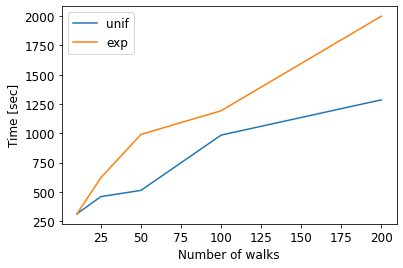}
\caption{Rule application time.}
\label{fig:application_time}
\end{figure}

\paragraph{Training and inference time}
The rule learning and rule application times are shown in the figures \ref{fig:learning_time} and \ref{fig:application_time}, dependent on the number of extracted temporal walks during learning.

The worst-case time complexity for learning rules of length $l$ is $\mathcal{O}(|\mathcal{R}|nlDb)$, where $n$ is the number of walks, $D$ the maximum node degree in the training set, and $b$ the number of body samples for estimating the confidence. 
The worst-case time complexity for inference is given by $\mathcal{O}(|\mathcal{G}| + |\mathcal{TR}_{r^q}|D^L|\mathcal{E}|\log(k))$, where $L$ is the maximum rule length in $\mathcal{TR}_{r^q}$ and $k$ the minimum number of candidates.
More detailed steps of the algorithms for understanding these complexity estimations are given by Algorithm \ref{alg:rule_learning_detailed} and Algorithm \ref{alg:rule_application_detailed}.

\begin{figure*}[h]
\centering
\includegraphics[width=1.8\columnwidth]{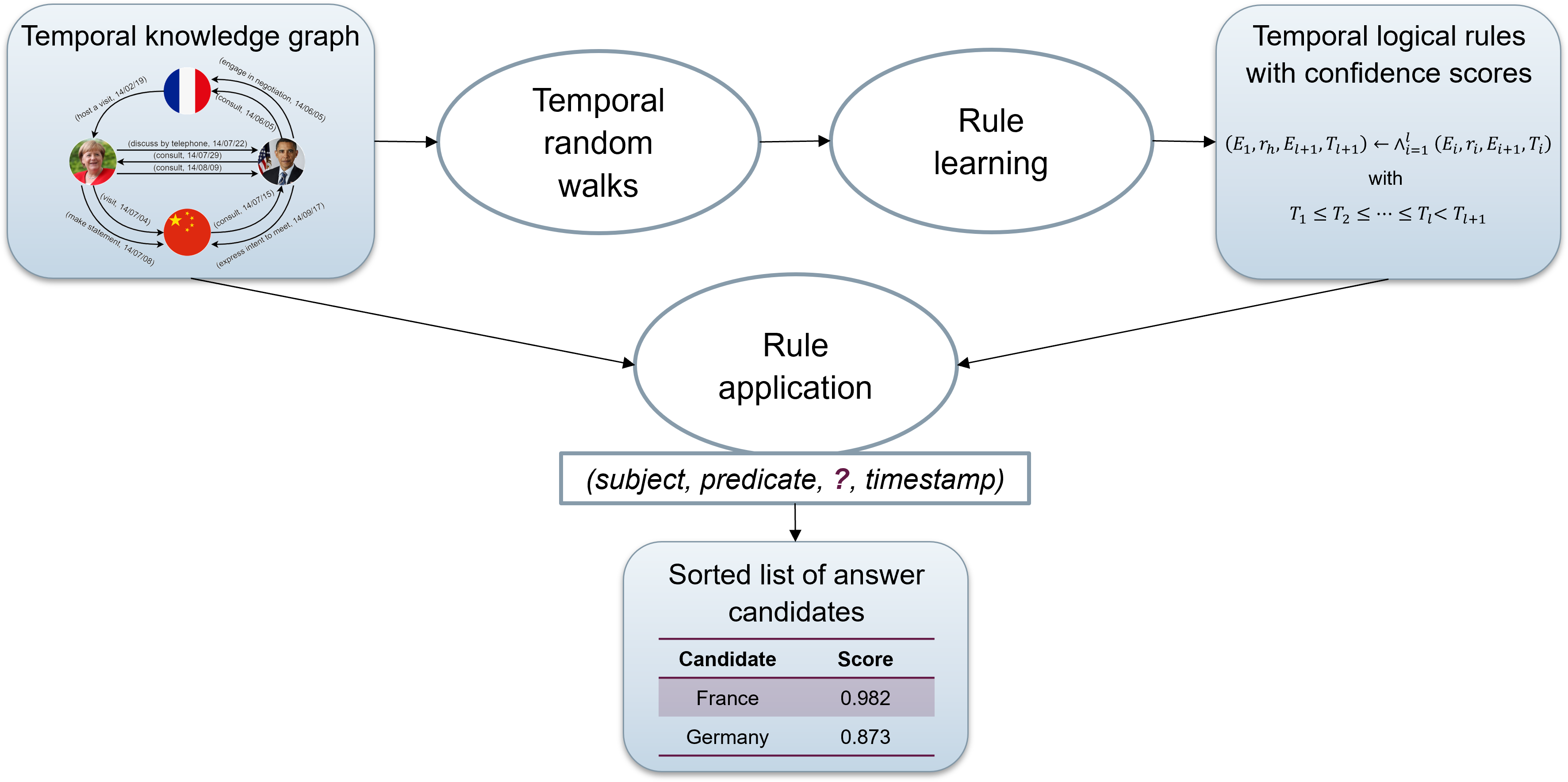}
\caption{Overall framework.}
\label{fig:framework}
\end{figure*}

\begin{algorithm*}[h]
\caption{Rule learning (detailed)}
\label{alg:rule_learning_detailed}
\textbf{Input}: Temporal knowledge graph $\mathcal{G}$.\\
\textbf{Parameters}: Rule lengths $\mathcal{L} \subset \mathbb{N}$, number of temporal random walks $n \in \mathbb{N}$, transition distribution $d \in \{\mathrm{unif}, \exp\}$.\\
\textbf{Output}:  Temporal logical rules $\mathcal{TR}$.

\begin{algorithmic}[1] 
\FOR{relation $r \in \mathcal{R}$} 
\FOR{$l \in \mathcal{L}$}
\FOR{$i \in [n]$}
\STATE{$\mathcal{TR}_r^l \leftarrow \emptyset$}
\STATE {\textbf{According to transition distribution $d$, sample a temporal random walk $W$ of length \mbox{$l+1$} with $t_{l+1} > t_l$.} 

\hfill $\triangleright$~See~\eqref{eq:temporal_walk}.\\
Sample uniformly a start edge $(e_s, r, e_o, t)$ with edge type $r$.}
\FOR{step $m \in \{2, \dots, l+1\}$}
\STATE{Retrieve adjacent edges of current object node.}
\IF{$m = 2$}
\STATE{Filter out all edges with timestamps greater than or equal to the current timestamp.}
\ELSE
\STATE{Filter out all edges with timestamps greater than the current timestamp.\\
Filter out the inverse edge of the previously sampled edge.}
\ENDIF
\IF{$m = l+1$}
\STATE{Retrieve all filtered edges that connect the current object to the source of the walk.}
\ENDIF
\STATE{Sample the next edge from the filtered edges according to distribution $d$.\\
\textbf{break} if there are no feasible edges because of temporal or cyclic constraints.}
\ENDFOR
\STATE{\textbf{Transform walk $W$ to the corresponding temporal logical rule $R$.} \hfill $\triangleright$~See~\eqref{eq:temporal_rule}.\\
Save information about the head relation and body relations.\\
Define variable constraints for recurring entities.}
\STATE{\textbf{Estimate the confidence of rule $R$.} \\
Sample $b$ body groundings. For each step $m \in \{1, \dots, l\}$, filter the edges for the correct body relation besides for the timestamps required to fulfill the temporal constraints.\\
For successful body groundings, check the variable constraints.\\
For each unique body, check if the rule head exists in the graph.\\
Calculate rule support / body support.}
\STATE{$\mathcal{TR}_r^l \leftarrow \mathcal{TR}_r^l \cup \{(R, \text{conf}(R))\}$}
\ENDFOR
\ENDFOR
\STATE{$\mathcal{TR}_r \leftarrow \cup_{l \in \mathcal{L}} \mathcal{TR}_r^l$}
\ENDFOR
\STATE{$\mathcal{TR} \leftarrow \cup_{r \in \mathcal{R}} \mathcal{TR}_r$}
\STATE \textbf{return} $\mathcal{TR}$
\end{algorithmic}
\end{algorithm*}

\begin{algorithm*}[h]
\caption{Rule application (detailed)}
\label{alg:rule_application_detailed}
\textbf{Input}: Test query $q = (e^q, r^q, ?, t^q)$, temporal logical rules $\mathcal{TR}$, temporal knowledge graph $\mathcal{G}$.\\
\textbf{Parameters}: Time window $w \in \mathbb{N} \cup \{\infty\}$, minimum number of candidates $k$, score function $f$.\\
\textbf{Output}: Answer candidates $\mathcal{C}$.

\begin{algorithmic}[1] 
\STATE{$\mathcal{C} \leftarrow \emptyset$

$\triangleright$ Apply the rules in $\mathcal{TR}$ by decreasing confidence.}
\STATE{\textbf{Retrieve subgraph $\mathcal{SG} \subset \mathcal{G}$ with timestamps $t \in [t^q-w, t^q)$.}\\
$\triangleright$ Only done if the timestamp changes. The queries in the test set are sorted by timestamp.\\
Retrieve edges with timestamps $t \in [t^q-w, t^q)$.\\
Store edges for each relation in a dictionary.}
\IF{$\mathcal{TR}_{r^q} \neq \emptyset$ }
\FOR{rule $R \in \mathcal{TR}_{r^q}$}
\STATE{\textbf{Find all body groundings of $R$ in $\mathcal{SG}$.}\\
Retrieve edges that could constitute walks that match the rule's body. First, retrieve edges whose subject matches $e^q$ and the relation the first relation in the rule body. Then, retrieve edges whose subject match one of the current targets and the relation the next relation in the rule body.\\
Generate complete walks by merging the edges on the same target-source entity.\\
Delete all walks that do not comply with the time constraints.\\
Check variable constraints, and delete the walks that do not comply with the variable constraints.}
\STATE{Retrieve candidates $\mathcal{C}(R)$ from the target entities of the walks.}
\FOR{$c \in \mathcal{C}(R)$}
\STATE{Calculate score $f(R,c)$. \hfill $\triangleright$ See~\eqref{eq:score_function}.}
\STATE{$\mathcal{C} \leftarrow \mathcal{C} \cup \{(c, f(R,c))\}$}
\ENDFOR
\IF{$|\{c \mid \exists R: (c, f(R,c)) \in \mathcal{C}\}| \geq k$}
\STATE{\textbf{break}}
\ENDIF
\ENDFOR
\ENDIF
\STATE \textbf{return} $\mathcal{C}$
\end{algorithmic}
\end{algorithm*}

\end{document}